# Analysis of Stability, Response and LQR Controller Design of a Small-Scale Helicopter Dynamics

**Hardian Reza Dharmayanda*, Taesam Kang**, Young Jae Lee***, and Sangkyung Sung******

Departement of Aerospace Information Engineering, Konkuk University
1 Hwayang-dong, Kwangjin-gu, Seoul, 143-701, Korea

*e-mail: king_arez@yahoo.com
**e-mail: tskang@konkuk.ac.kr
***e-mail: yjlee@konkuk.ac.kr
****e-mail: sksung@konkuk.ac.kr

### Abstract

This paper presents how to use feedback controller with helicopter dynamics state space model. A simplified analysis is presented for controller design using LQR of small scale helicopters for axial & forward flights. Our approach is simple and gives the basic understanding about how to develop controller for solving the stability of linear helicopter flight dynamics.

## 1 Introduction

Small scale helicopters have various applications which cannot be accomplished by humans due to emergency or dangerous situations such as an earthquake, flood, an active volcano, or a nuclear disaster [7]. They have many advantages compare to the fixed wing UAVs especially in their hovering ability. They also can be used to perform several difficult tasks during cruise. By those reasons, nowadays, small-scale helicopters become more popular for unmanned aerial vehicles. However miniature helicopters are more agile than full-scale helicopters because moments of inertia reduce as the fifth power of characteristic size, causing the miniature helicopters to have faster time responses. Since the main rotor tip speed of a small scale helicopter is equal to the tip speed of full-size helicopter, the ratio of the rotor moments to moments of inertia grows drastically as the size decreases. Then it is necessary to develop the model of helicopter flight dynamics and to design controller especially to overcome the agility problem of a small-scale helicopter in order to obtain stability condition for any sequence of flight i.e. taking off, hovering, and landing.

Furthermore, we can utilize this agile condition to execute more aggressive maneuvers which cannot be accomplished by full scale helicopter. The purpose of this work was to give the basic understanding about how to model and design controller of a small scaled helicopter flight dynamics. Thus we have restricted our research area into small-amplitude motions.

This paper will describe how to derive state space model from helicopter flight dynamics by linearizing the translational and rotational equations of motion. Some assumption has been made to reduce the complexity and this paper also is adapting small perturbation theory. Furthermore the model can be used for developing and evaluating controller design for axial and forward flight..

This paper will cover four main parts. First part describes the model of helicopter used in this work. Second part explains how to construct state-space model from identification of state parameters and control input. Third part gives the analysis of the system and feedback control simulation. The last part is the conclusion of this work.

## 2 Helicopter Model

For the purpose of simulating mathematical model and control system design, the Yamaha R-50 data is used [1]. The Yamaha R-50 helicopter uses a two-bladed main rotor with a Bell-Hiller stabilizer bar. The Bell-Hiller stabilizer bar is a secondary rotor consisting of a pair of paddles connected to the rotor shaft through an unrestrained teetering hinge. It receives the same cyclic control input as the main rotor do but it has a slower response than the main blades and is also sensitive to airspeed and wind gust.

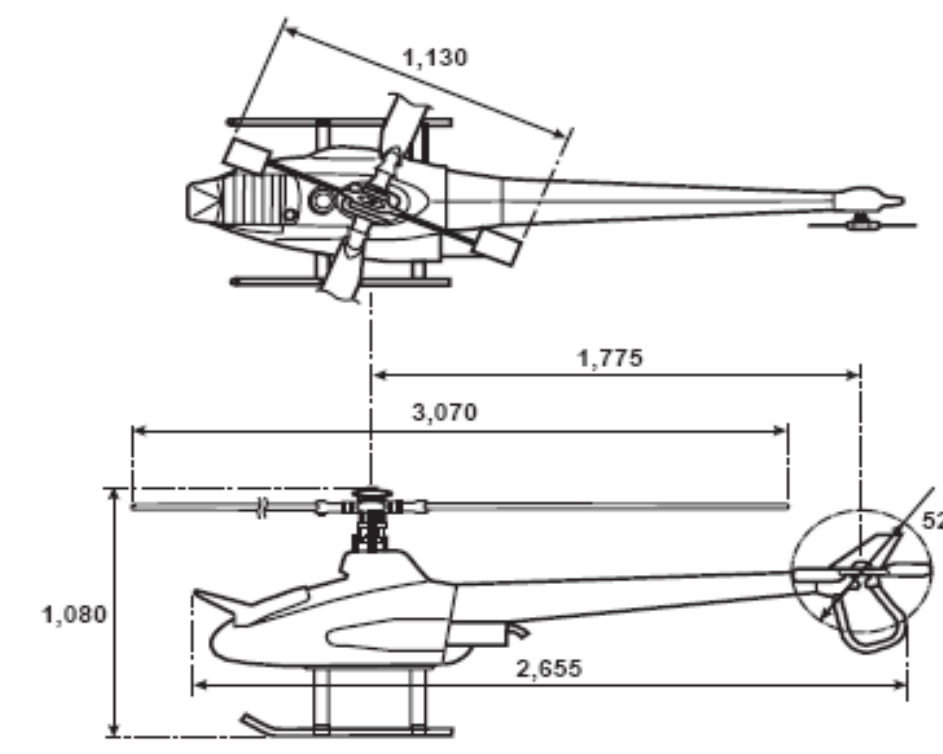

Figure 1 Small Scaled Helicopter

The specification of the R-50 is as follows:

Length: 3.5m
Width:0.7m
Height: 1.08m
Dry Weight: 44 kg
Payload: 20kg
Engine Output: 12 hp
Rotor Diameter: 3.070m
Flight time: 60 min
System operation time: 60 min

The reason why we choose this small helicopter model as a reference is that the model is already proven. Furthermore, the helicopter that we will use to control is quite similar to the R-50 model.

## 3 State Space Model Structure

State estimation is a fundamental need for autonomous vehicle. The accuracy that we demand from the estimation algorithm depends on the control system to be used [6]. There are six rigid body degrees of freedom; three translational (along the three body axes) and three rotational (about the same axes). The equations of helicopter dynamics are expressed with respect to the body axis coordinate system. The attitude of the body with respect to the inertial reference frame is defined by Euler angles: $\phi$ , $\theta$ , $\psi$ . The order of rotation is as follows: first, rotate along the X-axis by angle $\phi$ , then, along the Y-axis by angle $\theta$ , and finally along the Z-axis by angle $\psi$ . To each such unique order there exists a corresponding rotation matrix which transforms vectors in the body-centered frame to the inertial frame.

The basic equations of motion for a linear model of the helicopter dynamics are derived from the Newton-Euler equations for a rigid body that is free to rotate and translate simultaneously in all six degrees of freedom. The system matrix is obtained using lateral, longitudinal and rotational fuselage dynamics, stabilizer bar dynamics, and swash-plate actuator dynamics [1]. While applying the small-disturbance theory we assume that the motion of the helicopter consists of small deviations about a steady flight condition.

The equations of motion can be linearized using the small disturbance theory [2].

$$\dot{u} = v_0 r - w_0 q - g\theta + X_u u + X_a a \quad (1)$$
$$\dot{v} = w_0 r - u_0 q + g\phi + Y_v v + Y_b b \quad (2)$$
$$\dot{w} = u_0 q - v_0 p + Z_w w + Z_{col}\theta_{lc} \quad (3)$$
$$\dot{p} = L_u u + L_v v + L_b b \quad (4)$$
$$\dot{q} = M_u u + M_v v + M_a a \quad (5)$$
$$\dot{r} = N_r r + N_{ped}\theta_T - N_{ped} r_{fb} \quad (6)$$

Thus as we mentioned before, this theory cannot be applied to problems in which large-amplitude motions are to be expected. However, in many cases the small-disturbance theory yields sufficient accuracy for practical engineering purposes [4].

The rotor dynamics were represented by first order rotor flapping differential equations

$$\tau_f \dot{b} = -b - \tau_f p + B_a a + B_{lat}\theta_{ls} + B_{lon}\theta_{lc} \quad (7)$$
$$\tau_f \dot{a} = -a - \tau_f q + A_b b + A_{lat}\theta_{ls} + A_{lon}\theta_{lc} \quad (8)$$

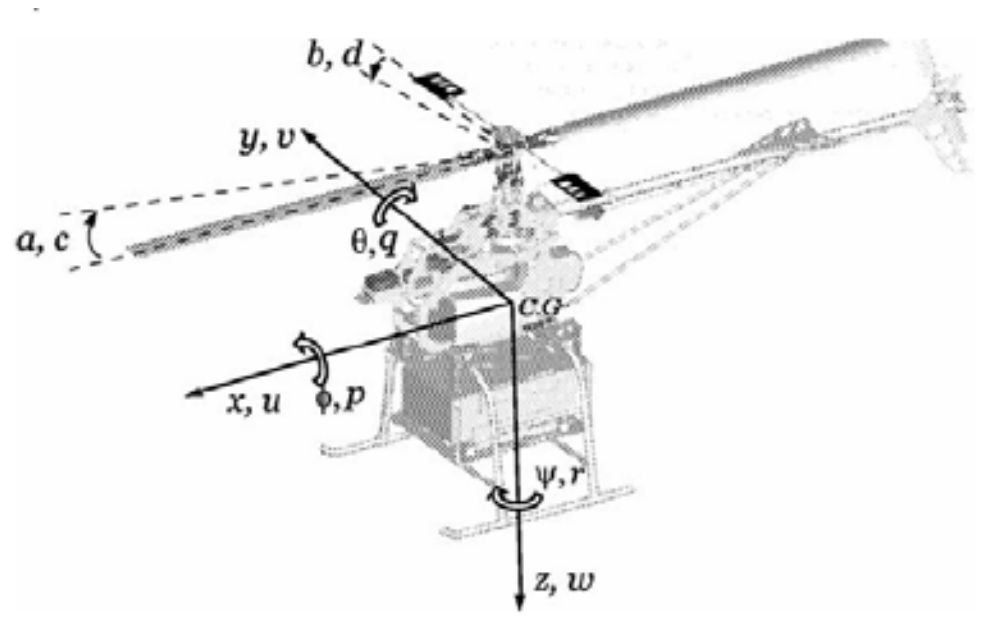

Figure 2 Coordinate definition

The stabilizer bar can be treated as secondary rotor. The blades receive the same cyclic inputs from the swash-plate similar to main rotor. We can develop the lateral and longitudinal stabilizer bar dynamic equation as:

$$\tau_s \dot{d} = -d - \tau_s p + D_{lat}\theta_{ls} \quad (9)$$
$$\tau_s \dot{c} = -c - \tau_s q + C_{lon}\theta_{lc} \quad (10)$$

The controller is designed using input-output linearization. In Helicopter control there are four control inputs available to the pilot. Those inputs are collective $(\theta_0)$, longitudinal cyclic $(\theta_{ls})$ , and lateral cyclic $(\theta_{lc})$ which are control inputs for the main rotor, and tail rotor collective $(\theta_{ot})$ , which is for the tail rotor. We can develop state vector as:

$$x = \begin{bmatrix} u & v & p & q & \phi & \theta & a & \mathrm{b} & \mathrm{w} & \mathrm{r} & \mathrm{r_{fb}} & \mathrm{c} & d \end{bmatrix}'$$

And the input parameters are

$$u = \begin{bmatrix} \theta_{ls} & \theta_{lc} & \theta_{ot} & \theta_{o} \end{bmatrix}'$$

Finally we construct the state space for flight dynamic of small scale helicopter as

$$\begin{bmatrix} \dot{u} \\ \dot{v} \\ \dot{p} \\ \dot{q} \\ \dot{\phi} \\ \dot{\theta} \\ \tau_f \dot{a} \\ \tau_f \dot{b} \\ \dot{w} \\ \dot{r} \\ \dot{r}_{fb} \\ \dot{c} \\ \dot{d} \end{bmatrix} = \begin{bmatrix} X_u & 0 & 0 & 0 & 0 & -g & X_a & 0 & 0 & 0 & 0 & 0 & 0 \\ 0 & Y_v & 0 & 0 & g & 0 & Y_b & 0 & 0 & 0 & 0 & 0 & 0 \\ L_u & L_v & 0 & 0 & 0 & 0 & 0 & L_b & L_w & 0 & 0 & 0 & 0 \\ M_u & M_v & 0 & 0 & 0 & 0 & M_a & 0 & M_w & 0 & 0 & 0 & 0 \\ 0 & 0 & 1 & 0 & 0 & 0 & 0 & 0 & 0 & 0 & 0 & 0 & 0 \\ 0 & 0 & 0 & 1 & 0 & 0 & 0 & 0 & 0 & 0 & 0 & 0 & 0 \\ 0 & 0 & 0 & -\tau_f & 0 & 0 & -1 & A_b & 0 & 0 & 0 & A_c & 0 \\ 0 & 0 & -\tau_f & 0 & 0 & 0 & B_a & -1 & 0 & 0 & 0 & 0 & B_d \\ 0 & 0 & 0 & 0 & 0 & 0 & Z_a & Z_b & Z_w & Z_r & 0 & 0 & 0 \\ 0 & N_v & N_p & 0 & 0 & 0 & 0 & 0 & N_w & N_r & N_{rfb} & 0 & 0 \\ 0 & 0 & 0 & 0 & 0 & 0 & 0 & 0 & 0 & K_r & K_{rfb} & 0 & 0 \\ 0 & 0 & 0 & -\tau_s & 0 & 0 & 0 & 0 & 0 & 0 & 0 & -1 & 0 \\ 0 & 0 & -\tau_s & 0 & 0 & 0 & 0 & 0 & 0 & 0 & 0 & 0 & -1 \end{bmatrix} \begin{bmatrix} u \\ v \\ p \\ q \\ \phi \\ \theta \\ a \\ b \\ w \\ r \\ r_{fb} \\ c \\ d \end{bmatrix} + \begin{bmatrix} 0 & 0 & 0 & 0 \\ 0 & 0 & Y_{ped} & 0 \\ 0 & 0 & 0 & 0 \\ 0 & 0 & 0 & M_{col} \\ 0 & 0 & 0 & 0 \\ 0 & 0 & 0 & 0 \\ A_{lat} & A_{lon} & 0 & 0 \\ B_{lat} & B_{lon} & 0 & 0 \\ 0 & 0 & 0 & Z_{col} \\ 0 & 0 & N_{ped} & N_{col} \\ 0 & 0 & 0 & 0 \\ 0 & C_{lon} & 0 & 0 \\ D_{lat} & 0 & 0 & 0 \end{bmatrix} \begin{bmatrix} \theta_{ls} \\ \theta_{lc} \\ \theta_{t} \\ \theta_{0} \end{bmatrix} \quad (11)$$

For simplicity we neglect the heading angle because it has no effect on Aerodynamic Forces and Moments of Helicopter.

*Table 1 Nomenclatures

| Symbols | Details |
|---|---|
| Φ | Euler roll angle |
| θ | Euler pitch angle |
| u, v, w | translational speed in x, y, z body axis |
| uo, vo, wo | trim condition for translational speed |
| p, q, r | angular rate in x, y, z body plane axis |
| Xa, Yb | rotor forces derivatives |
| Ma, Lb | flapping spring derivatives |
| Xu, Yu, Lu, Lv, Mu, Mv | general aerodynamic effects are expressed by speed derivatives |
| τs | stabilizer's bar time constant |
| d | lateral stabilizer bar dynamic |
| c | longitudinal stabilizer bar dynamic |
| Dlat, Clon | stabilizer's bar input derivatives |
| τs | main rotor time constant |
| Blat, Blon, Alat, Alon | main rotor input derivatives |
| Ba, Ab | rotor cross coupling effect |
| Nped | sensitivity to pedal control |
| Nr | bar airframe damping coefficient |
| rfb | low pass filter |
| Krfb | low pass filter pole |

## 4 Flight Dynamics Simulation and Feedback Controller

Here we represent a simulation for forward flight motion. In order to know the characteristics of dynamics system we can check the eigenvalues of the system or we could check the system response. Identified parameters were used in this dynamic model consists of helicopter and flight data stated in [1]. We will see that the helicopter is not stable but it is controllable thus we could make it stable by applying feedback control. We applied 3 degree of collective pitch for 5 seconds to see the open loop response. The simulation shows clearly that the system is unstable and it produces diverging response Fig. 3.

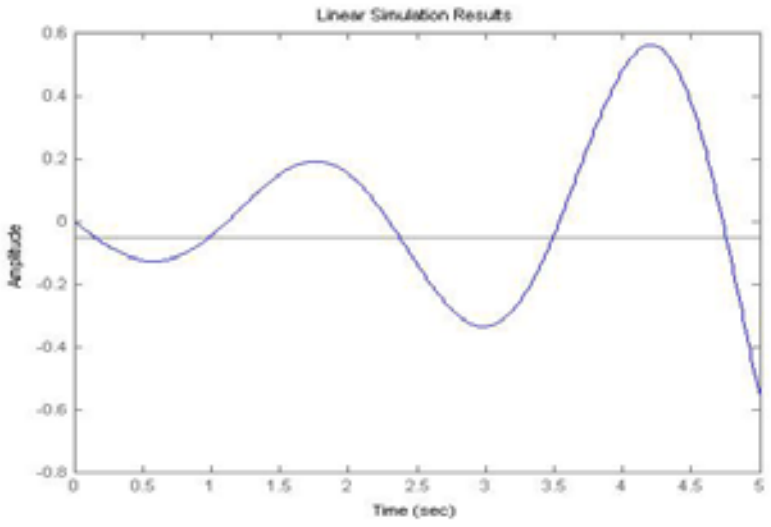

Fig. 3 Open loop response for forward flight

If we check eigenvalues we can see that we have positive real parts which indicate that the system is unstable.

*Table 2 Eigenvalues of the System

| | | | |
|---|---|---|---|
| | | **7** | 0.42+2.62i |
| **1** | -0.12 | **8** | 0.42-2.62i |
| **2** | 0 | **9** | 0 |
| **3** | -5.85+7.34i | **10** | -0.16 |
| **4** | -5.85-7.34i | **11** | -3.36 |
| **5** | 0.68+3.55i | **12** | -2.83 |
| **6** | 0.68-3.55i | **13** | -1.01 |

*for forward flight

One of the methods is using feedback controller where we must find gain matrix so that we could obtain the stable eigenvalues as desired. The gain matrix will be obtained using the method of linear quadratic (LQR) state-feedback regulator for continuous plant. After using this gain for controller we will obtain convergence response Fig. 4.

In this work, Q is a identity matrix constructed from the output matrix of the system and R is an identity matrix developed from the input matrix of the system. It is necessary to find the best weighting factor so that the controller gets the system act accordingly. Q matrix is define as

$$Q=100\begin{bmatrix} 9\times10^{6} & 0 & 0 & 0 & 0 & 0 & 0 & 0 & 0 & 0 & 0 & 0 & 0 \\ 0 & 4\times10^{2} & 0 & 0 & 0 & 0 & 0 & 0 & 0 & 0 & 0 & 0 & 0 \\ 0 & 0 & 9\times10^{5} & 0 & 0 & 0 & 0 & 0 & 0 & 0 & 0 & 0 & 0 \\ 0 & 0 & 0 & 2.8\times10^{6} & 0 & 0 & 0 & 0 & 0 & 0 & 0 & 0 & 0 \\ 0 & 0 & 0 & 0 & 9\times10^{5} & 0 & 0 & 0 & 0 & 0 & 0 & 0 & 0 \\ 0 & 0 & 0 & 0 & 0 & 10^{-3} & 0 & 0 & 0 & 0 & 0 & 0 & 0 \\ 0 & 0 & 0 & 0 & 0 & 0 & 10^{-3} & 0 & 0 & 0 & 0 & 0 & 0 \\ 0 & 0 & 0 & 0 & 0 & 0 & 0 & 10^{6} & 0 & 0 & 0 & 0 & 0 \\ 0 & 0 & 0 & 0 & 0 & 0 & 0 & 0 & 10^{-3} & 0 & 0 & 0 & 0 \\ 0 & 0 & 0 & 0 & 0 & 0 & 0 & 0 & 0 & 10^{-3} & 0 & 0 & 0 \\ 0 & 0 & 0 & 0 & 0 & 0 & 0 & 0 & 0 & 0 & 0 & 0 & 0 \\ 0 & 0 & 0 & 0 & 0 & 0 & 0 & 0 & 0 & 0 & 0 & 0 & 0 \\ 0 & 0 & 0 & 0 & 0 & 0 & 0 & 0 & 0 & 0 & 0 & 0 & 0 \end{bmatrix}$$

*for forward flight

While R matrix used identity matrix 4 by 4.

Table 3 Table of Identified Paramters

| Parameters | Axial Flight | Forward Flight |
|---|---|---|
| tauf | 0.04631 | 0.03463 |
| hcg | -0.4109 | -0.3212 |
| taus | 0.3415 | 0.2591 |
| Xu | -0.05046 | -0.1217 |
| Xa | -32.2 | -32.2 |
| Xr | 0 | -11 |
| Yv | -0.1539 | -0.1551 |
| Yb | 32.2 | 32.2 |
| Yr | 0 | -49.2 |
| Lu | -0.1437 | 0 |
| Lv | 0.1432 | 0 |
| Lw | 0 | -0.2131 |
| Lb | 166.1 | 213.2 |
| Mu | -0.05611 | 0 |
| Mv | -0.0585 | 0 |
| Mw | 0 | 0.07284 |
| Ma | 82.57 | 108 |
| Ba | 0.3681 | 0.4194 |
| Bd | 0.7103 | 0.6638 |
| Ab | -0.1892 | -0.1761 |
| Ac | 0.6439 | 0.5773 |
| Zb | -131.2 | 0 |
| Za | -9.748 | 0 |
| Zw | -0.6141 | -1.011 |
| Zr | 0.9303 | 0 |
| Zp | 0 | 11 |
| Zq | 0 | 49.2 |
| Np | -3.525 | 0 |
| Nv | 0.03013 | 0.4013 |
| Nw | 0.08568 | 0 |
| Nr | -4.129 | -3.897 |
| Nrfb | -33.07 | -26.43 |
| Kr | 2.163 | 2.181 |
| Krfb | -8.258 | -7.794 |
| Blat | 0.1398 | 0.1237 |
| Blon | 0.0138 | 0.02003 |
| Alat | 0.03127 | 0.02654 |
| Alon | -0.1004 | -0.08372 |
| Zcol | -45.84 | -60.27 |
| Mcol | 0 | 6.98 |
| Ncol | -3.329 | 0 |
| Nped | 33.07 | 26.43 |
| Dlat | 0.2731 | 0.2899 |
| Clon | -0.2587 | -0.225 |
| Yped | 0 | 11.23 |
| tauped | 0.0991 | 0.05893 |

*Table 4 Gain Matrix using LQR

| K matrix | column 1 | column 2 | column 3 |
|---|---|---|---|
| Row 1 | -6191.54 | 0.0231 | 9283.79 |
| Row 2 | 27694.62 | 0.0039 | 2076.93 |
| Row 3 | 0.1738 | 199.97 | -1.111 |
| Row 4 | -2821.57 | -0 | 4.71 |

| column 4 | column 5 | column 6 | column 7 |
|---|---|---|---|
| 2255.62 | 9258.21 | 10019.96 | 1272.24 |
| -10116.26 | 2070.16 | -44889.21 | -6703.54 |
| -0.0629 | -0.7648 | -0.2801 | -0.0211 |
| 2372.29 | 4.7131 | 9882.52 | 6.92 |

| column 8 | column 9 | column 10 |
|---|---|---|
| 5510.49 | 261.22 | -0 |
| 1463.76 | -1171.58 | -0 |
| 2.0971 | -0.0073 | 0.0006 |
| 1.2021 | -9911.29 | 0 |

| column 11 | column 12 | column 13 |
|---|---|---|
| -0 | 1.062 | 5.1039 |
| -0 | -6.56 | 1.6179 |
| 0.0014 | 0 | 0 |
| 0 | 0 | -0 |

*for forward flight

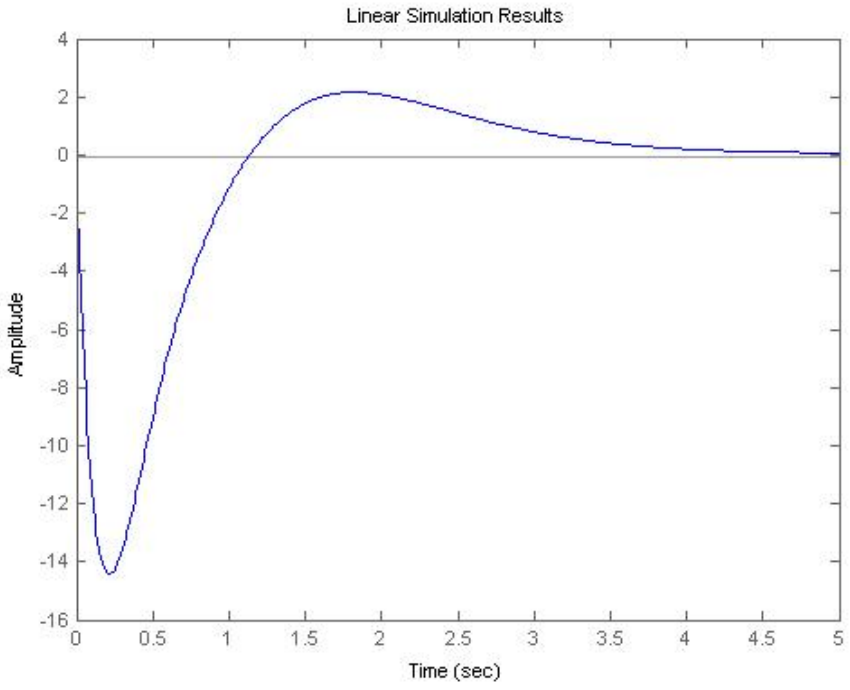

Figure 4 Close Loop Response

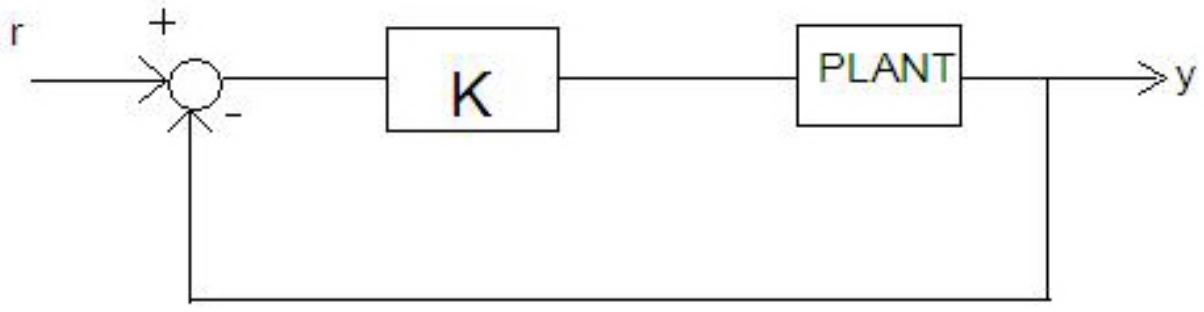

Figure 5 General Block Diagram

First, we develop simulation using Matlab Simulink which simulated block diagram in fig. 5.

Plant was considered as state space model with disturbance by some measurement noise (fig. 6). Measurement noise is added to simulate realistic situation. By using controller gain K we want to minimize the error so that the small helicopter obey the input command (fig. 7)

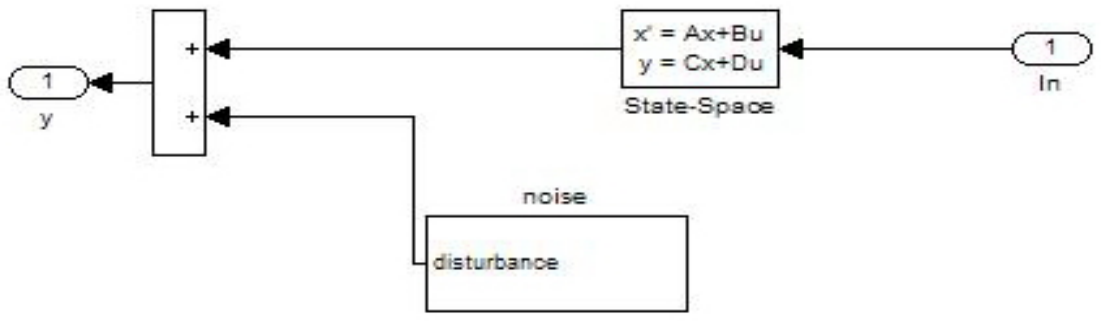

Figure 6 Plant Block Diagram

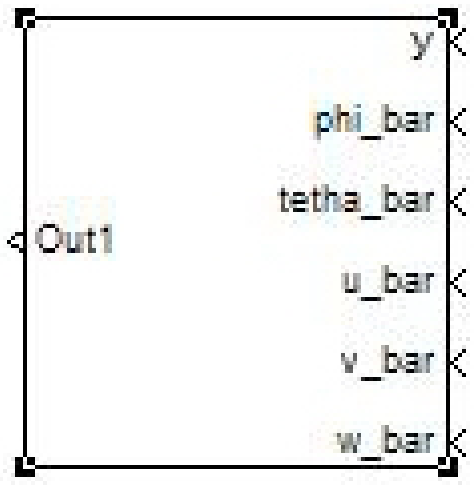

Figure 7 Input Command Block

Now, let’s check how the controller works by observing angles, translational and angular velocity.

For longitudinal forward flight the small helicopter has been commanded to flight with several conditions:

- Forward speed u = 10 ft/s
- Side speed and axial speed is kept zero
- The attitudes of small helicopter are 3 deg of pitch and 1.5 deg of roll

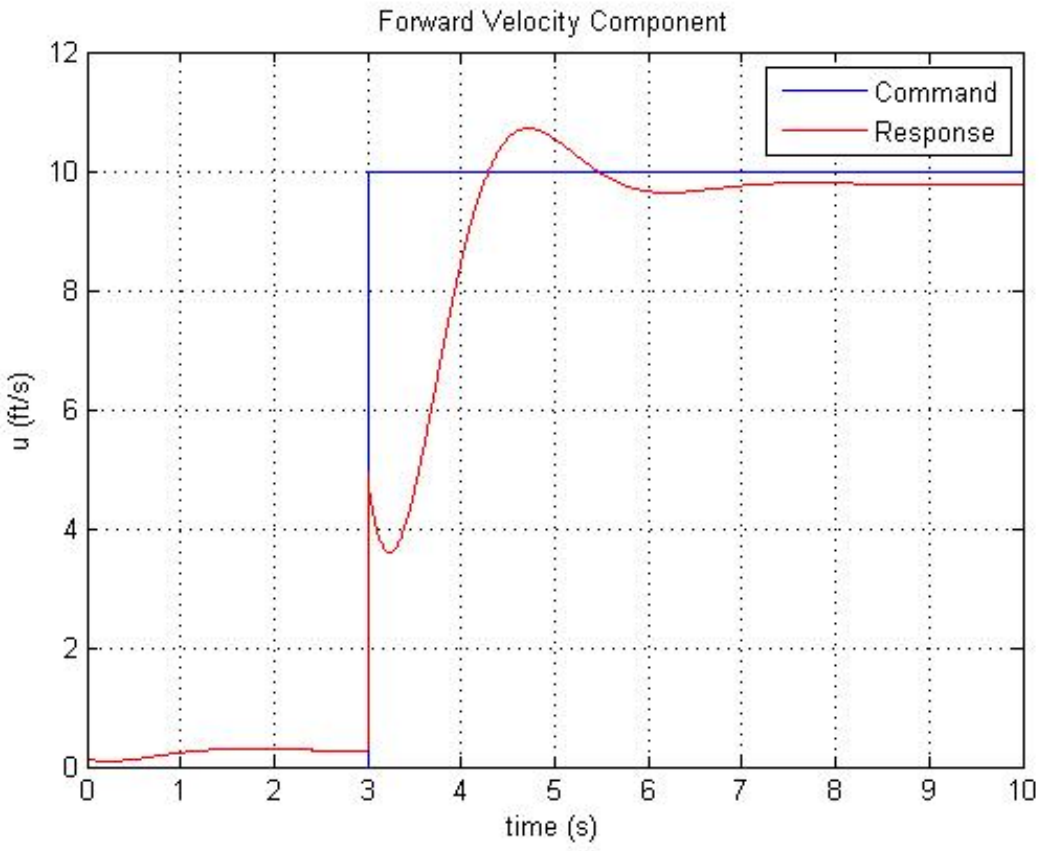

Figure 8 Forward Velocity Response

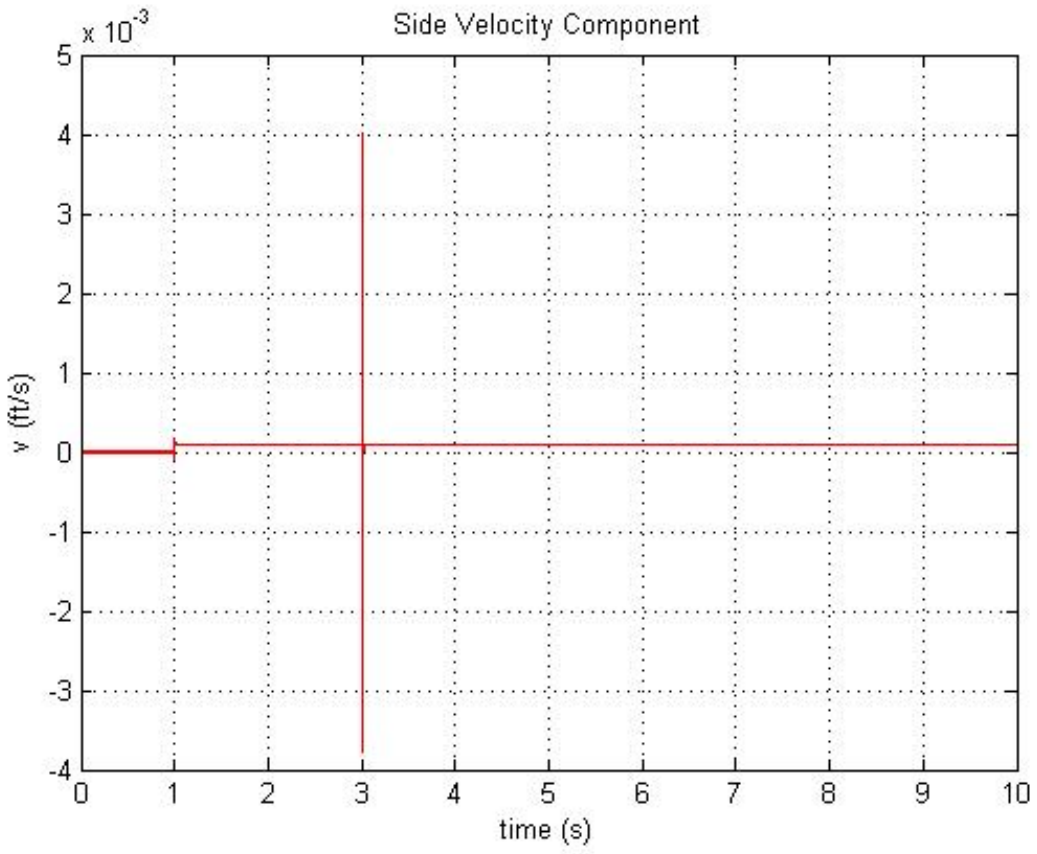

Figure 9 Side Velocity Response

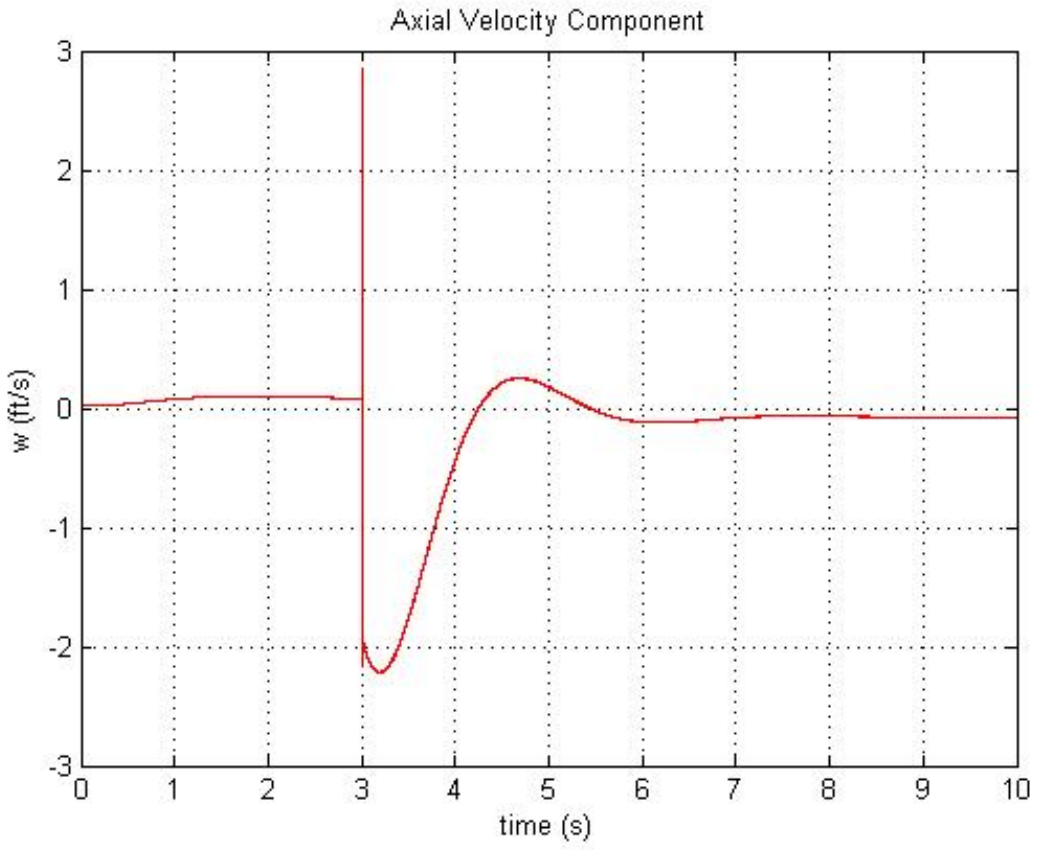

Figure 10 Axial Velocity Response

As we can see from fig. 8 – 10 for translational velocity, the controller achieved forward velocity of 10ft/s while it kept another component of translational velocity zero for any input.

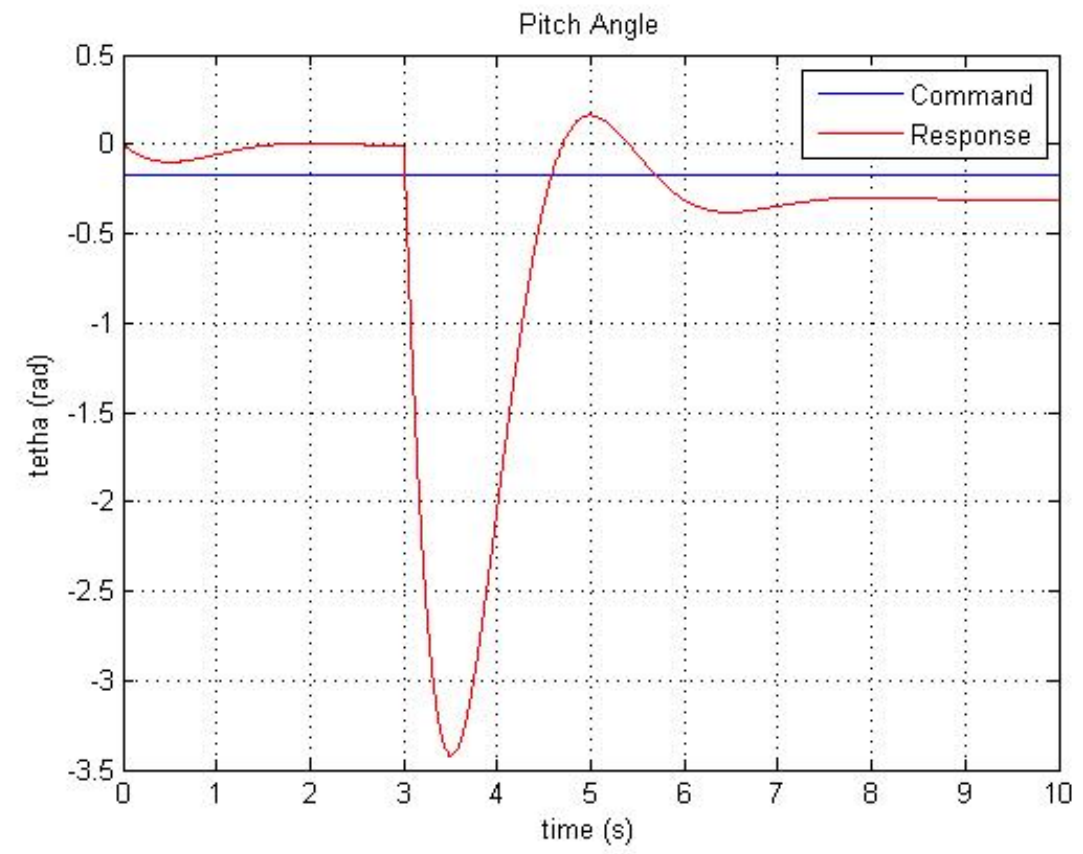

Figure 11 Pitch Angle Response

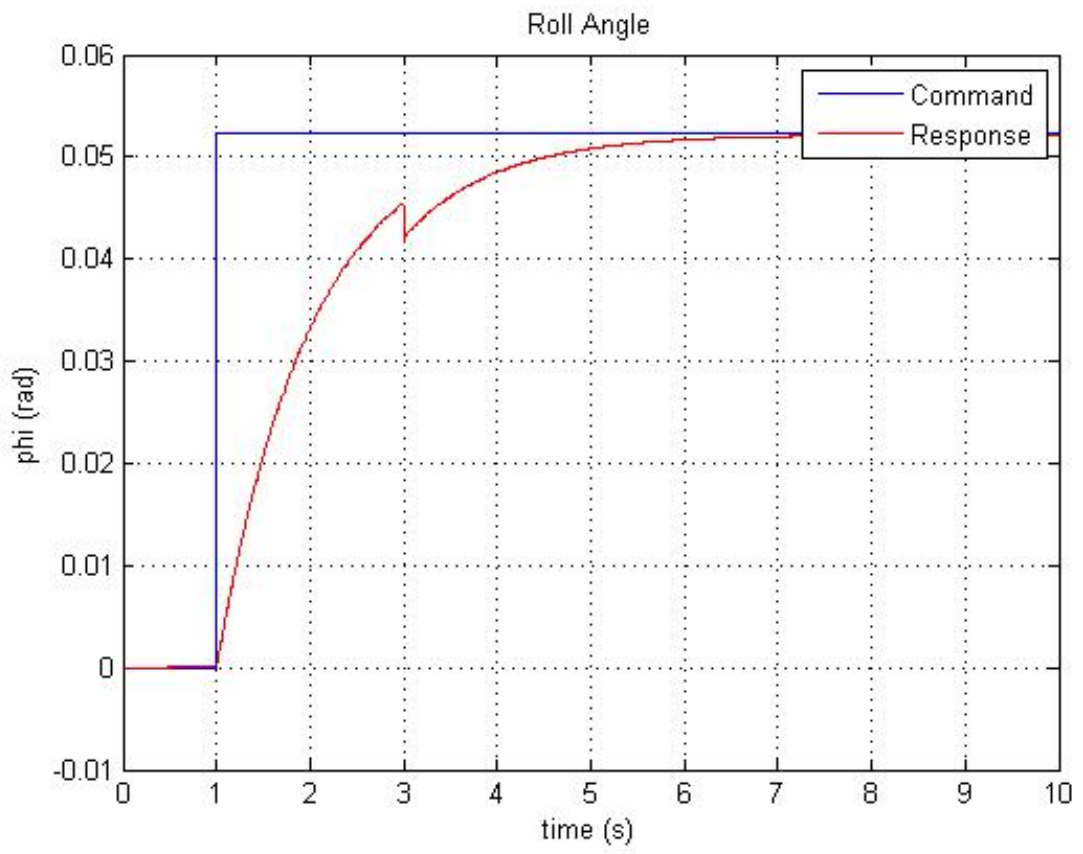

Figure 12 Roll Angle Response

We mentioned before that we want the helicopter attitude meet some criteria which were the pitch angle 3 deg and roll angle 1.5 deg. The controller worked well by accommodating those input command.

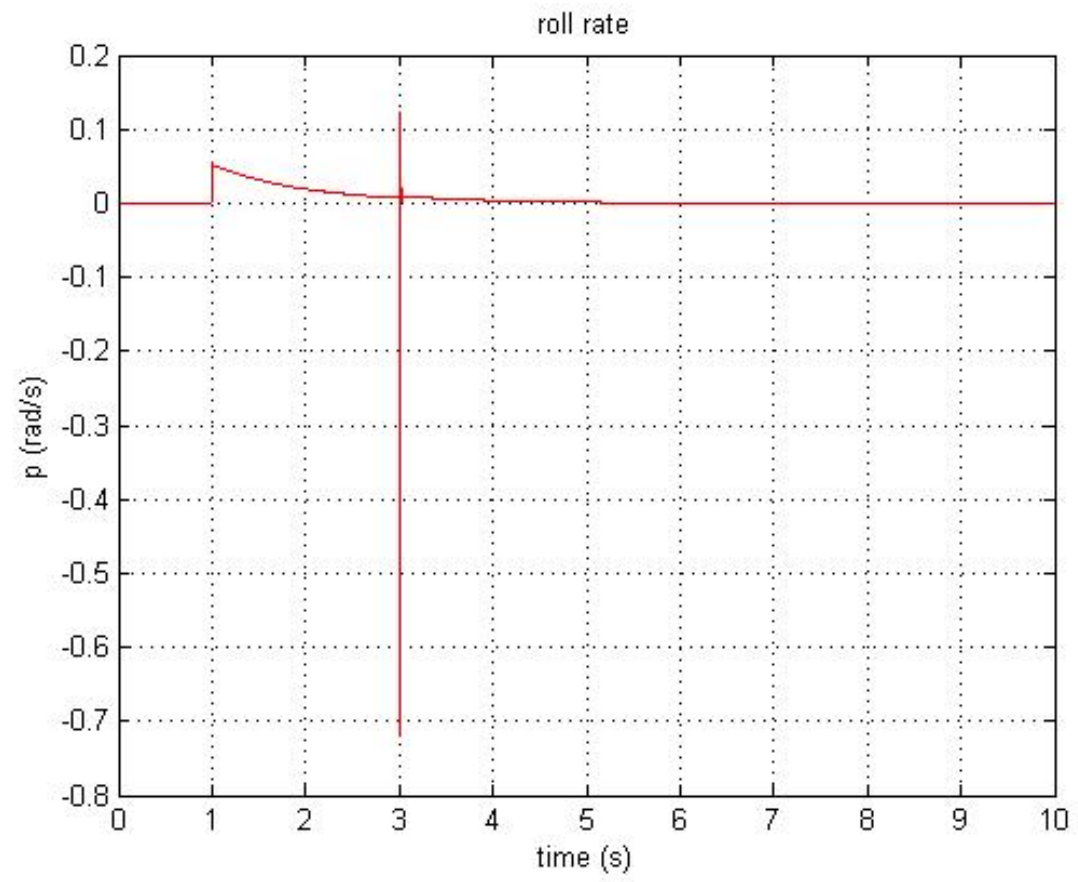

Figure 13 Roll Rate Response

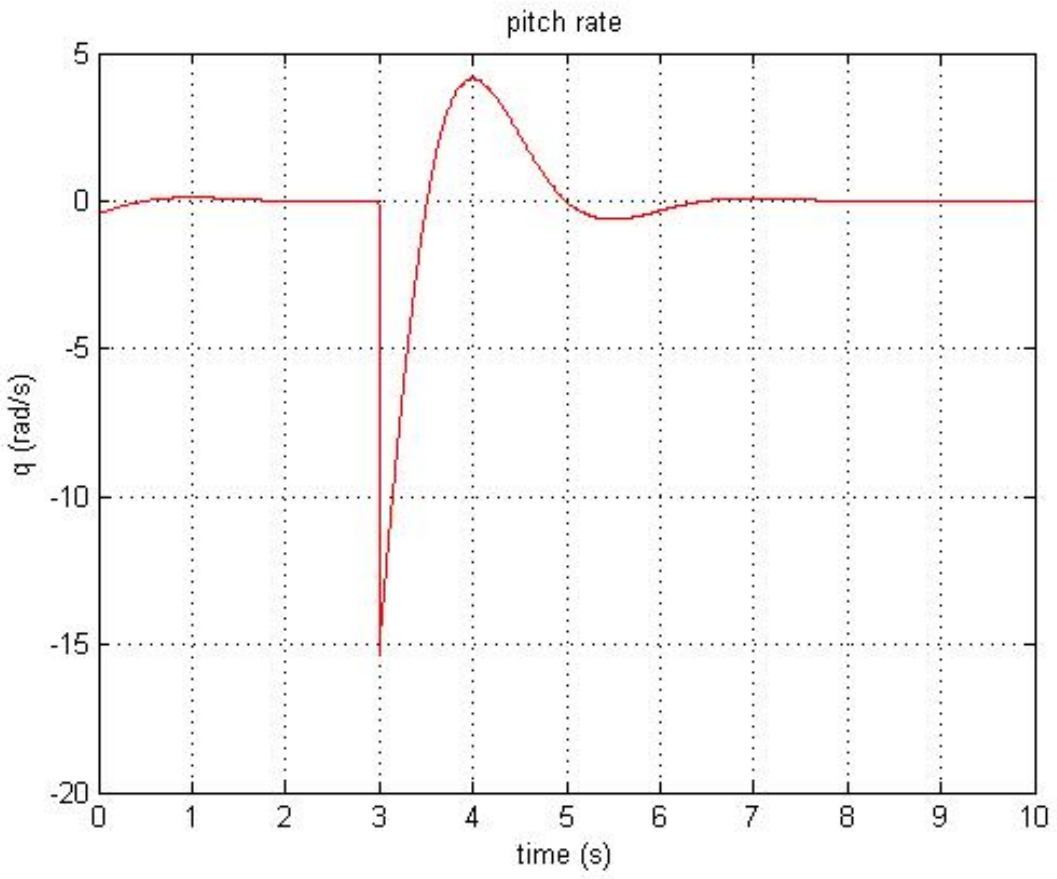

Figure 14 Pitch Rate Response

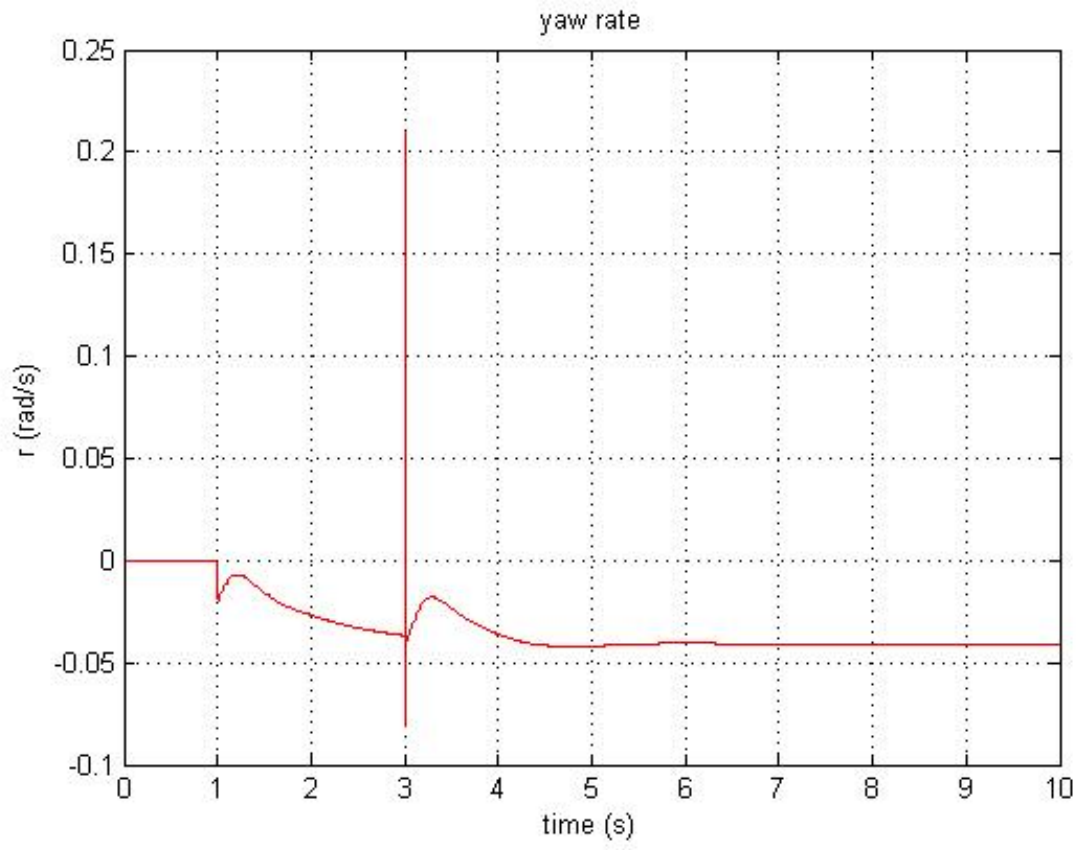

Figure 15 Yaw Rate Response

Finally, it is obvious from those figures above that the controller gives the stability for angular velocity as well.
The stability of the small helicopter was achieved by keeping the angular rates small.

For axial flight, we maintain all the criterions above to be zero except for the axial velocity. The vertical climb velocity 10ft/s was commanded.

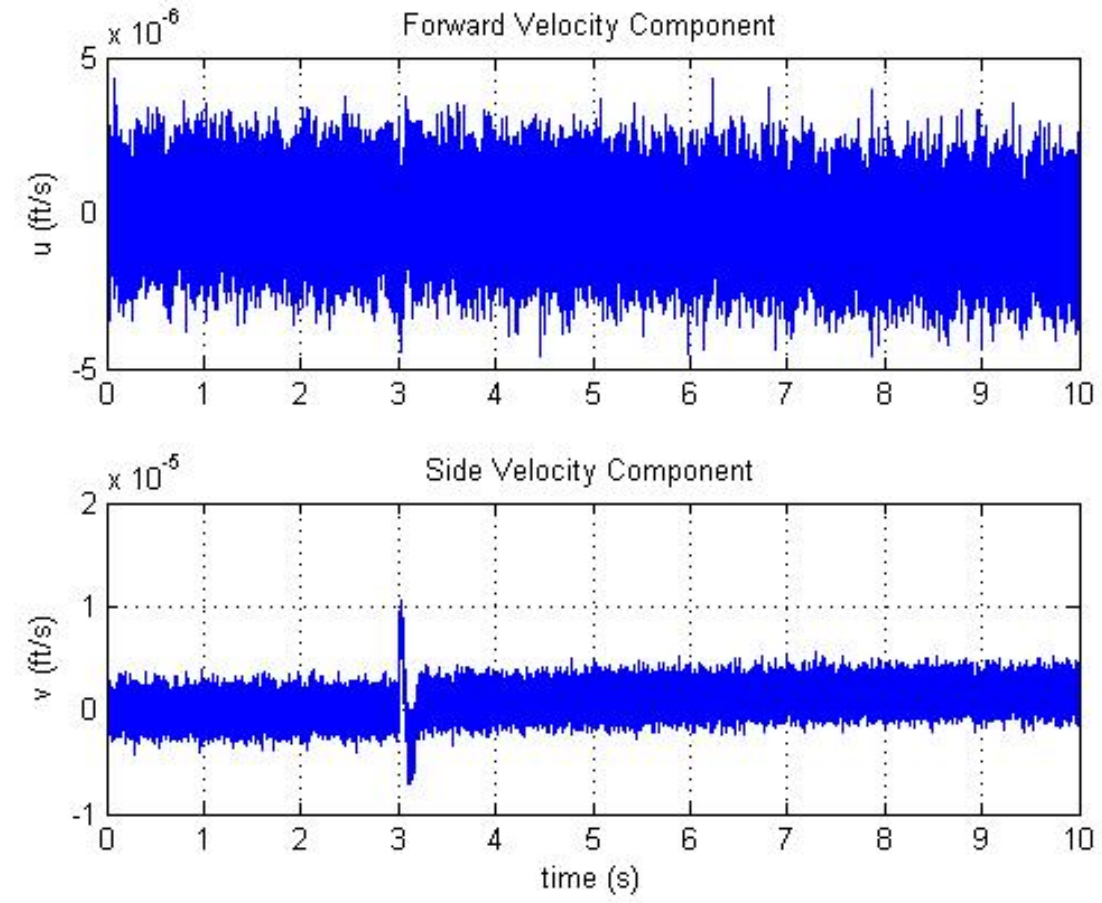

Figure 16 Translational Velocity for Axial Flight

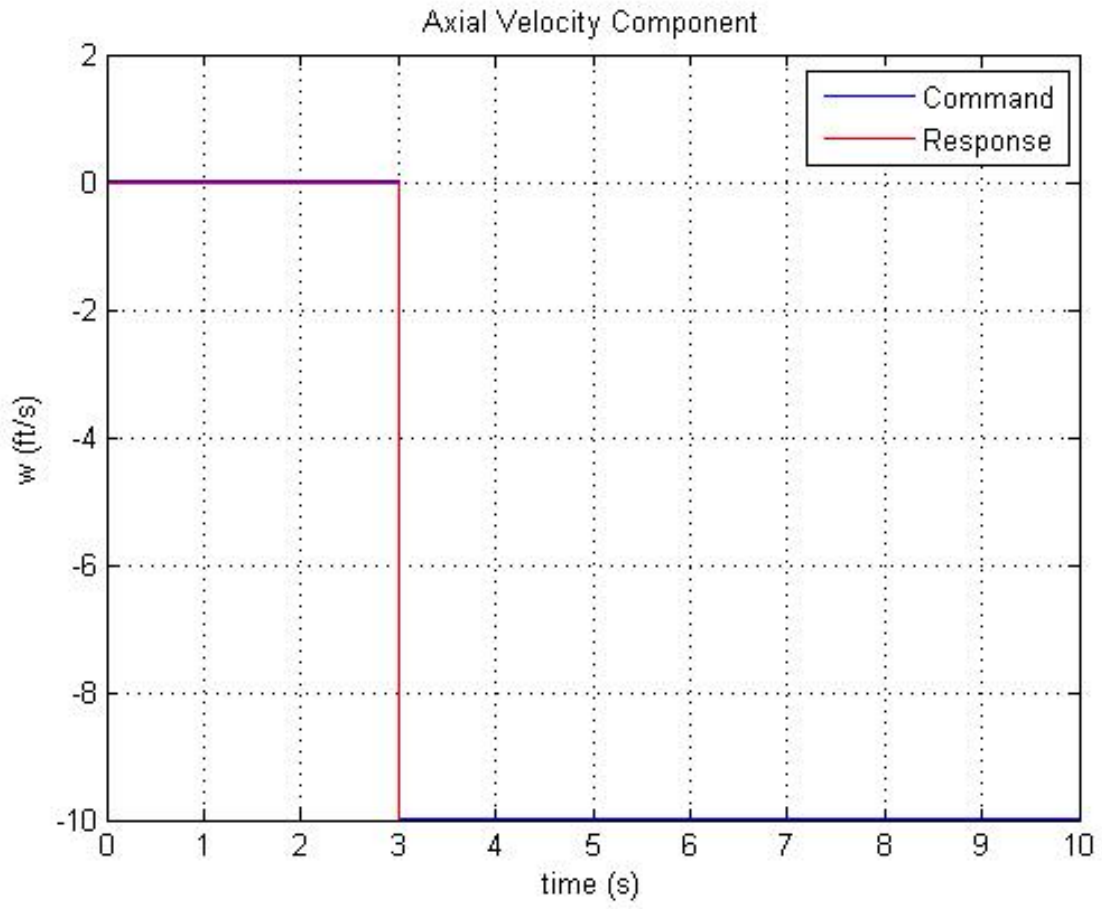

Figure 17 Vertical Climb Response

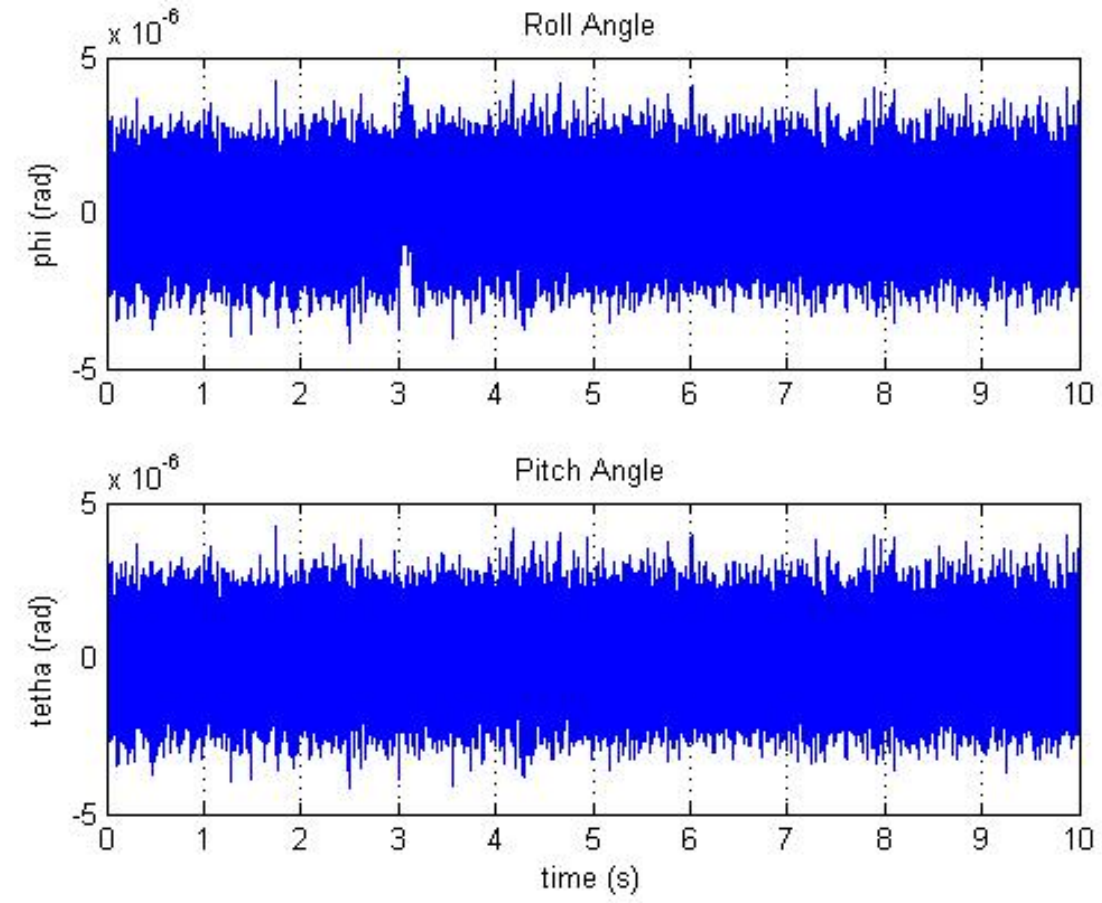

Figure 18 Attitude Response for Axial Flight

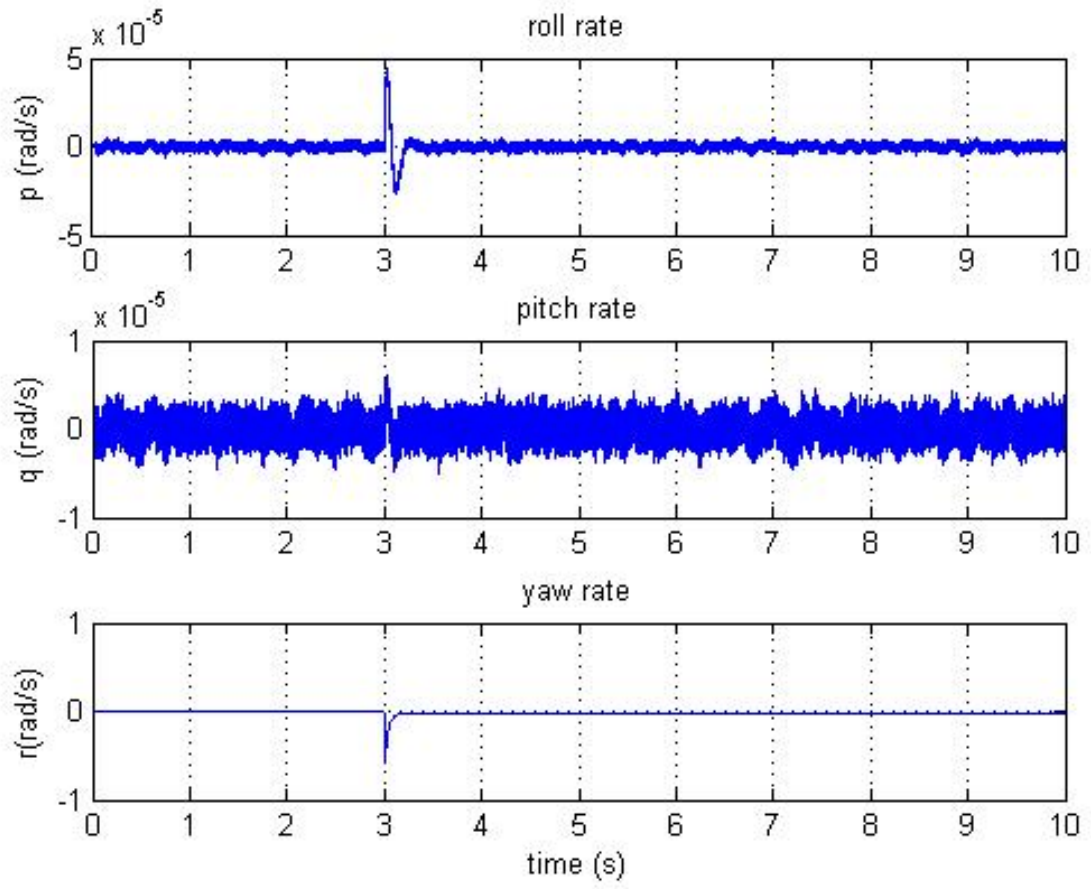

Figure 19 Angular Rate Response for Axial Flight

This controller design offers the advantage of simpler stability analysis so that it doesn't cost burden for computer. Furthermore, It give the basic understanding about how to construct controller for small scale helicopter flight dynamic. It is useful because it intrinsically deal multi-input multi-output systems effectively.

## 5 Conclusions

In this paper we used state space dynamic models of a small-scaled helicopter, to check its stability and constructed stabilizing control gain matrix using LQR method. This paper also shows us the simulation using feedback controller. Future works is intended to build autonomous control using H-infinity controller which uses several state measurements only.

## 6 Acknowledgement

This work was supported by the Korea Research Foundation Grant Funded by the Korea Government (MOEHRD) (KRF-2006-211-D00026) & BK 21